\documentclass[runningheads]{llncs}

\usepackage[T1]{fontenc}
\usepackage{graphicx}
\usepackage{amsfonts}
\usepackage{amsmath}
\usepackage{booktabs}
\usepackage{subcaption}
\usepackage{float}

\begin{document}

\title{Predicting Response to Neoadjuvant Chemotherapy in Ovarian Cancer from CT Baseline Using Multi-Loss Deep Learning}
\titlerunning{Predicting Response to Neoadjuvant Chemotherapy}

\author{Francesco Pastori*\inst{1} \and
Francesca Fati*\inst{1,2} \and
Marina Rosanu\inst{1} \and
Luigi De Vitis\inst{3} \and
Lucia Ribero\inst{1} \and       
Gabriella Schivardi\inst{1} \and
Giovanni Damiano Aletti\inst{1,4} \and
Nicoletta Colombo\inst{1} \and
Jvan Casarin\inst{5} \and
Francesco Multinu\inst{1,4} \and
Elena De Momi\inst{1,2}
}

\authorrunning{F. Pastori et al.}

\institute{Department of Gynecologic Oncology, European Institute of Oncology, IEO, IRCCS, Milan, Italy \and
Department of Electronics, Information and Bioengineering, Politecnico di Milano, Milan, Italy \and
Department of Obstetrics and Gynecology, Mayo Clinic, Rochester, USA \and
Department of Oncology and Hemato-Oncology, University of Milan, Milan, Italy \and
Department of Medicine and Innovative Technology, Università degli Studi dell'Insubria, Varese, Italy \\
Corresponding author: \email{francesco.pastori@mail.polimi.it}
}

\maketitle

\begin{abstract}
Ovarian cancer is the most lethal gynecologic malignancy: around 60\% of patients are diagnosed at an advanced stage, with an associated 5-year survival rate of about 30\%. Early identification of non-responders to neoadjuvant chemotherapy remains a key unmet need, as it could prevent ineffective therapy and avoid delays in optimal surgical management. \\
This work proposes a non-invasive deep learning framework to predict neoadjuvant chemotherapy response from pre-treatment contrast-enhanced CT by leveraging automatically derived 3D lesion masks.
The approach encodes axial slices with a partially fine-tuned pretrained image encoder and aggregates slice-level representations into a volumetric embedding through an attention-based module. Training combines classification loss with supervised contrastive regularization and hard-negative mining to improve separation between ambiguous responders and non-responders. The method was developed on a retrospective single-center cohort from the European Institute of Oncology (Milan, IT), including 280 eligible patients (147 responder, 133 non-responder). On the test cohort, the model achieved a ROC-AUC of 0.73 (95\% CI: 0.58–0.86) and an F1-score of 0.70 (95\% CI: 0.56–0.82).
Overall, these results suggest that the proposed architecture learns clinically relevant predictive patterns and provides a robust foundation for an imaging-based stratification tool.

\end{abstract}

\section{Introduction}
\label{sec:introduction}
Ovarian cancer is the most lethal gynecologic malignancy,
60\% of patients are diagnosed when the tumor is in an advanced stage and they present a 5-year survival rate of 30\% \cite{stat4}. The epithelial ovarian cancer accounts for about 90\% of all cases and its most prevalent and aggressive form is the high-grade serous ovarian carcinoma, which constitutes the majority of advanced-stage diagnoses and is the primary focus of this work \cite{intro0}. Standard management typically involves primary debulking surgery aiming for macroscopic complete resection of the disseminated disease, followed by platinum-based adjuvant chemotherapy.
When complete resection is unlikely due to extensive disease burden, or when patients cannot undergo upfront surgery due to clinical condition, an alternative pathway is neoadjuvant chemotherapy which is commonly administered as three cycles of carboplatin and paclitaxel, followed by interval debulking surgery and additional postoperative chemotherapy. \cite{intro8}

A key unmet need is that, even with careful preoperative evaluation, approximately 40\% of patients assigned to neoadjuvant chemotherapy fail to achieve any objective benefit from carboplatin paclitaxel therapy and would likely benefit from alternative strategies if identified earlier as non-responders \cite{crispin2023integrated}. 
Given the aggressive nature of ovarian cancer, time is a critical factor: predicting, before treatment initiation, whether a tumor will respond to neoadjuvant chemotherapy could prevent ineffective therapy and avoid delays in definitive surgical management. Addressing this challenge requires a predictive tool that is rapid, non-invasive, and widely available in routine clinical practice.

Building on this clinical need, this work proposes a deep learning based predictive framework that leverages pre-treatment contrast-enhanced CT scans to classify patients as responders or non-responders to neoadjuvant chemotherapy, with the aim of supporting treatment planning and enabling more personalized and effective management of ovarian cancer.

\section{State of the Art}
\label{sec:State of the Art}

Recent advances in artificial intelligence for medical imaging, especially deep learning, have demonstrated the ability to extract complex imaging patterns that may be imperceptible to visual inspection, motivating radiomics and deep learning approaches for treatment response prediction from pre-treatment scans.

Radiomics extracts handcrafted texture and shape descriptors from segmented tumor regions of interest and applies feature selection to reduce redundancy; results can depend on segmentation quality and imaging variability.
Crispin-Ortuzar et al.\cite{crispin2023integrated} developed an integrated radiogenomic model combining clinical and CT-derived radiomic features to predict neoadjuvant chemotherapy (NACT) response in ovarian cancer. Using an ensemble of Elastic Net, SVM, and Random Forest trained on 72 patients, the model achieved an AUC of 0.78, substantially outperforming the clinical-only baseline (AUC 0.47), highlighting the added value of radiomics.

Deep learning enables end-to-end representation learning directly from images. A key benchmark for this work is the multicenter study by Yin et al.\cite{yin2023predicting}, which predicted NACT response from pre-treatment CT combined with clinical features, using RECIST 1.1 as reference standard for response evaluation and including 757 patients. They developed a 2D slice-based convolutional ResNet architecture achieving an external AUC of 0.82 for response prediction.

In contrast to convolutional architectures operating on single 2D slices, emerging state-of-the-art approaches employ Vision Transformers (ViT) to model long-range dependencies, analyze complete 3D volumes, and integrate contrastive learning to strengthen representation learning.
Fati et al.\cite{fati2025deep} introduced OVIT, a ViT-based model that encodes CT slices and aggregates them with attention to predict resectability. OVIT was trained on 465 patients, achieving a ROC-AUC of 0.80 on the test set and suggesting that ViT-based models can serve as clinically relevant decision support systems.

Gupta et al.\cite{gupta2024margin} trained a ResNet-50 with a margin-aware contrastive loss for histopathology image classification across 5 public datasets (more than 100k images), reporting an AUC of 0.96 on binary breast histopathology classification, concluding that adding a margin constraint improves representation quality and generalization.

Building on these studies, this work jointly exploits full volumetric lesion information, Vision Transformers, and supervised contrastive learning to provide a practical foundation for imaging-based decision support, enabling early stratification of NACT responders and non-responders and potentially reducing ineffective therapy while supporting more timely, personalized treatment planning.

\section{Methods}
\label{sec:sec_materials_and_methods}
\subsection{Model Architecture}

\begin{figure}[t]
    \centering
    \makebox[\textwidth][c]{%
        \includegraphics[width=1.4\textwidth]{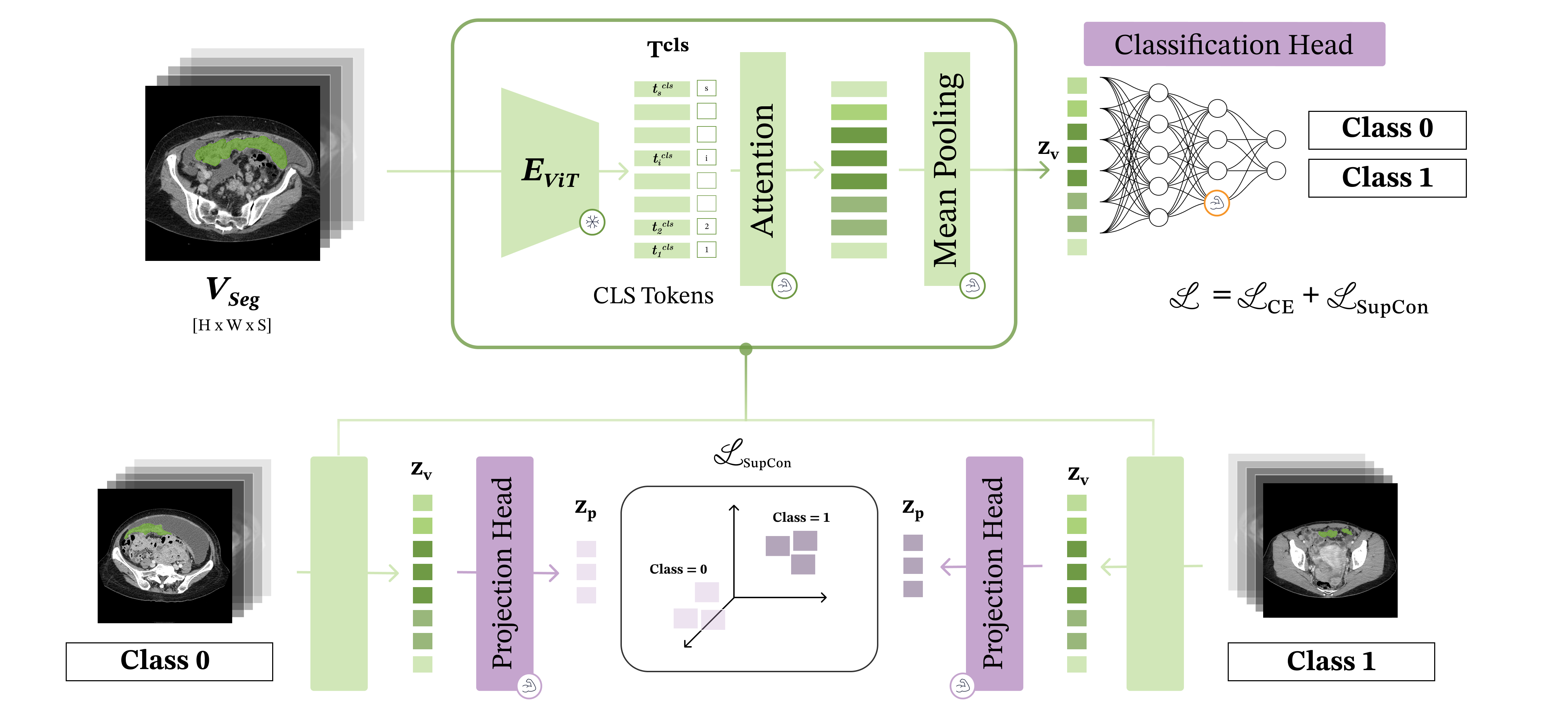}
    }
    \caption{Overview of the proposed model architecture for NACT response prediction. Pre-treatment 3D CT volumes are processed by a segmentation model to obtain lesion mask volumes, which are reshaped into axial slices. Each slice is encoded independently by a pretrained encoder. Slice-level representations are aggregated by an attention module and average pooling to produce a single volume embedding. This embedding is shared by two MLP heads: a classification head to predict non-responder(0) vs. responder(1), and a projection head to obtain embeddings optimized with supervised contrastive learning. Training uses a multi-loss objective combining supervised contrastive and cross-entropy loss.}
    \label{fig:model_architecture}
\end{figure}

An overview of the proposed pipeline is reported in Figure~\ref{fig:model_architecture}. The model predicts response to NACT from 3D pre-treatment CT scans $\mathbf{V_{CT}} \in \mathbb{R}^{H \times W \times S \times 1}$, where $H$, $W$ and $S$ denote height, width and number of axial slices, respectively. The network outputs a binary prediction $\hat{y} \in \{0,1\}$ where $0$ denotes a non-responder and $1$ a responder, via a model $M$:
\begin{equation}
    \hat{y} = M(\mathbf{V_{CT}}).
\end{equation}
From the raw $\mathbf{V}_{CT}$, a lesion mask volume is obtained through an open source deep learning segmentation model $M_S$:
\begin{equation}
    \mathbf{V}_{seg} = M_S(\mathbf{V}_{CT}),
    \qquad
    \mathbf{V}_{seg} \in \mathbb{R}^{H \times W \times S \times 1}
\end{equation}

To leverage a 2D pretrained ViT encoder, the input volume is reshaped into a sequence of axial slices $\mathbf{V_{seg,i}} \in \mathbb{R}^{H \times W \times 1}$.
Each slice is converted to a 3-channel representation and independently processed by a pretrained ViT encoder $E_{\mathrm{ViT}}$.

For each slice $i$, the encoder produces token embeddings:
\begin{equation}
    \mathbf{T}_i = E_{\mathrm{ViT}}(\mathbf{V_{seg,i}}),
    \qquad
    \mathbf{T}_i \in \mathbb{R}^{(N+1)\times d}
\end{equation}
where $N$ is the number of patch tokens and $d$ is the embedding dimension.

The corresponding classification token $\mathbf{t}^{\mathrm{cls}}_i \in \mathbb{R}^{d}$ is extracted as a global descriptor of the slice. The classification tokens from all $S$ slices are concatenated to form a compact representation of the 3D input in the ViT feature space:
\begin{equation}
    \mathbf{T}^{\mathrm{cls}} =
    [
        \mathbf{t}^{\mathrm{cls}}_{1};
        \dots;
        \mathbf{t}^{\mathrm{cls}}_{S}
    ],
    \qquad
    \mathbf{T}^{\mathrm{cls}} \in \mathbb{R}^{S \times d}
\end{equation}

A learnable positional embedding along the slice axis is added to $\mathbf{T}^{\mathrm{cls}}$ to preserve ordering information. The resulting sequence is aggregated by an attention pooling module that models inter-slice relationships.
The attention output is aggregated via mean pooling along the slice dimension, yielding a single embedding vector per input volume.
The combination of slice-level ViT feature extraction, attention pooling, and mean aggregation is denoted as the volume-encoding module $M_V$:
\begin{equation}
\mathbf{z_V} = M_V(\mathbf{V_{seg}}),
\qquad
\mathbf{z_V} \in \mathbb{R}^{d}.
\end{equation}

From the shared volume embedding $\mathbf{z_V}$, the network branches into two heads: the classification head and the projection head.
The classification head $H_{\mathrm{cls}}$ is implemented as an MLP composed of multiple linear layers interleaved with Layer Normalization, GELU non-linear activations, and dropout regularization. Starting from the pooled embedding dimension $d$, the head progressively reduces dimensionality through two hidden transformations ($d \rightarrow d/2 \rightarrow d/8$) and outputs the final 2-class logits ${\hat{y}} \in \mathbb{R}^{2}$ ($d/8 \rightarrow 2$)
\begin{equation}
\hat{y} = H_{\mathrm{cls}}(\mathbf{z_V}),
\qquad \hat{y} \in \mathbb{R}^{2},
\end{equation}

The logits are converted to class probabilities via softmax, the corresponding class label is then obtained by setting a decision threshold on the predicted probability.

In parallel, the projection head $H_{\mathrm{proj}}$ maps the same embedding $\mathbf{z_V}$ to a low-dimensional representation $\mathbf{z_p}$ used for supervised contrastive learning:
\[
\mathbf{z_p} = H_{\mathrm{proj}}(\mathbf{z_V}), \qquad \mathbf{z_p} \in \mathbb{R}^{p}.
\]
In the proposed model, $H_{\mathrm{proj}}$ is a shallow MLP with intermediate Layer Normalization and GELU activation. The projection is subsequently L2-normalized to obtain an embedding $\mathbf{\tilde{z}_p}$ on the unit hypersphere:
\[
\mathbf{{\tilde{z}_p}} = \frac{\mathbf{z_p}}{\lVert \mathbf{z_p} \rVert_2}, \qquad \mathbf{\tilde{z}_p} \in \mathbb{R}^{p}.
\]

\subsection{Training Strategies}
A transfer learning framework was adopted, in which a pre-trained ViT encoder was fine-tuned and two task-specific heads were trained from scratch: a classification head for binary NACT response prediction and a projection head for contrastive embedding learning.
Training is performed with a multi-loss function that combines a cross-entropy classification loss and a supervised contrastive loss:
\begin{equation}
\label{multiloss}
\mathcal{L} = \mathcal{L}_{\mathrm{CE}}(\hat{y},y) + \alpha\ \cdot \mathcal{L}_{\mathrm{SupCon}}(\mathbf{\tilde{z}_p}, y),
\end{equation}
where $y \in \{0,1\}$ is the ground-truth response label and $\alpha$ (from 0 to 0.3) controls the relative contribution of the contrastive objective.

For the supervised contrastive loss, we consider pairs of volumes $(\mathbf{V_{seg}}_1, \mathbf{V_{seg}}_2)$ with a supervised pair label $s \in \{0,1\}$, where $s=1$ denotes a positive pair (same ground truth labels) and $s=0$ a negative pair (different ground truth labels). Let $\mathbf{\tilde{z}_{p_1}}, \mathbf{\tilde{z}_{p_2}} \in \mathbb{R}^{p}$ be the embeddings produced by the projection head and normalized, and let $D = \lVert \mathbf{\tilde{z}_{p_1}} - \mathbf{\tilde{z}_{p_2}} \rVert_2$ be the Euclidean distance between the two embeddings. The contrastive loss is:
\[
\mathcal{L}_{\mathrm{SupCon}}(D,s) = s \cdot D^2 + (1-s) \cdot [\max(0, m-D)]^2,
\]
where $m > 0$ is the margin. Intuitively, the loss reduces the distance between positive pairs (pulling them toward $D \approx 0$) and enforces that negative pairs are separated by at least a margin $m$, penalizing only negatives that are ``too close'' ($D < m$).

The multi-loss equation (\ref{multiloss}) is a weighted combination where $\alpha$ controls the contribute of the contrastive component. During training, $\alpha$ is increased linearly from 0 up to a maximum value $\alpha_{\max}$ (0.3), so as cross-entropy provides an initial strong signal to organize the representations, while the progressive increase of $\alpha$ allows the contrastive term to act as a regularizer of the embedding space.
At each iteration, the classification component is optimized on individual volumes using cross-entropy, while the contrastive component is optimized on volume pairs constructed from the labels. Pairs for contrastive learning are generated using two complementary strategies. In the initial epochs, both positive and negative pairs are formed by randomly sampling examples from the training set, providing a stable and low-variance contrastive signal while the representation space is still being structured. In a later stage, hard mining is enabled to build negative pairs by selecting samples with different labels that lie close to each other in the embedding space (hard negatives). This focuses optimization on the most ambiguous and informative cases, yielding a more effective contrastive gradient than “easy” negative pairs that are already well separated. In practice, the model is progressively encouraged to distinguish tumors that appear highly similar but exhibit different responses to NACT, improving its ability to capture subtle imaging cues associated with NACT response.

\section{Experiments}
\subsection{Dataset}

The study population was provided by the European Institute of Oncology (IEO) and included patients with histologically confirmed ovarian cancer, treated with NACT from 2016 to 2023.
All patients had available contrast-enhanced portal venous phase CT scans at diagnosis and after three NACT cycles. RECIST 1.1 \cite{eisenhauer2009new} criteria is used to establish target label classifying a patient as responder if the post-treatment tumor burden decreases by at least 30\% compared with baseline, otherwise non-responder.
After applying quality and eligibility criteria, 280 patients (147 responder, 133 non-responder) were included. Of these, 226 were used for model development, while an independent test set of 54 patients was reserved for final evaluation.
For each patient all cancer lesions were segmented using an open source deep learning model OvSeg\cite{buddenkotte2023deep}.

\subsection{Implementation Details}
The pretrained ViT encoder selected for the experiments is DINOv3 Base Patch16 \cite{simeoni2025dinov3} thanks to its strong semantic prior.

Fine-tuning of the pretrained encoder was performed in a parameter-efficient manner using LoRA \cite{hu2022lora} adapters, restricting adaptation to the last Transformer block while keeping the remaining backbone frozen. This strategy updates only a small subset of additional parameters, reducing computational cost and limiting the risk of overfitting, which is particularly important in the presence of small dataset.
All experiments were trained on a single NVIDIA A100-PCIE GPU with 40 GB of memory. Optimization was performed using AdamW for up to 100 epochs with early stopping based on validation performance. The learning rate was set to $1 \times 10^{-4}$ and decreased following a cosine-shaped schedule, to regularize the classifier we applied a dropout rate of 0.2, the contrastive term $\alpha$ grows linearly up to $\alpha_{\max}$.

\subsection{Performance metrics}

The response prediction performance was evaluated using receiver operating characteristic (ROC) analysis, reporting the area under the curve (\text{ROC-AUC}). The ROC curve plots the true positive rate (\text{TPR}) against the false positive rate (\text{FPR}) as the threshold varies:
{\small
\begin{equation*}
\begin{aligned}
\mathrm{TPR} &= \frac{\mathrm{TP}}{\mathrm{TP}+\mathrm{FN}}, \qquad
\mathrm{FPR} = \frac{\mathrm{FP}}{\mathrm{FP}+\mathrm{TN}}.
\end{aligned}
\end{equation*}
\begin{equation*}
\mathrm{ROC\text{-}AUC} = \int_{0}^{1} \mathrm{TPR}(\mathrm{FPR}) \, d\mathrm{FPR}.
\end{equation*}}
In addition, we reported accuracy, precision, recall, and F1:
{\small
\begin{equation*}
\begin{aligned}
\mathrm{Acc} = \frac{\mathrm{TP}+\mathrm{TN}}{\mathrm{TP}+\mathrm{TN}+\mathrm{FP}+\mathrm{FN}}, \qquad
\mathrm{Recall} = \mathrm{TPR},
\end{aligned}
\end{equation*}
\begin{equation*}
\begin{aligned}
\mathrm{Precision} = \frac{\mathrm{TP}}{\mathrm{TP}+\mathrm{FP}}, \qquad
\mathrm{F1} = \frac{2\,\mathrm{Precision}\,\mathrm{Recall}}{\mathrm{Precision}+\mathrm{Recall}}.
\end{aligned}
\end{equation*}}
The classification threshold was selected on the validation set by maximizing F1, and the same threshold was then used for test evaluation.
To further characterize the model, we performed an attention-weight analysis, a test-set calibration assessment, and a confidence-aware error analysis.
Statistical significance versus the baseline was assessed using DeLong’s test for ROC-AUC, and a paired stratified bootstrap for accuracy, precision, recall, and F1 (p < 0.05).

\section{Results}
\begin{table}[t]
\centering
\scriptsize
\setlength{\tabcolsep}{3pt}
\renewcommand{\arraystretch}{1.12}
\caption{Quantitative performance comparison between configurations: each table entry features the median value with 95\% CIs. For each metric, the best value is in bold. Asterisks (*) indicate statistically significant differences versus the baseline (p<0.05).}
\label{tab:table_performance}

\makebox[\textwidth][c]{%
\resizebox{1.15\textwidth}{!}{%
\begin{tabular}{l|cccc}
\toprule
\textbf{Metric} & \textbf{Yin et al.} & \textbf{MultiLoss-Frozen} & \textbf{CrossEntropy-FineTuned} & \textbf{MultiLoss-FineTuned} \\
\midrule
\multicolumn{5}{l}{\textit{Confusion Matrix Elements (point estimates, 95\% CI)}} \\
\midrule
True Positives [$\uparrow$] & 20 (13--27) & \textbf{31 (24--38)} & 15 (9--22) & 21 (14--28) \\
False Positives [$\downarrow$] & 12 (6--18) & 15 (9--22) & \textbf{3 (0--7)} & 5 (1--10) \\
True Negatives [$\uparrow$] & 8 (3--14) & 5 (1--9) & \textbf{17 (10--24)} & 15 (9--22) \\
False Negatives [$\downarrow$] & 14 (8--21) & \textbf{3 (0--7)} & 19 (12--26) & 13 (7--19) \\
\midrule
\multicolumn{5}{l}{\textit{Classification Metrics (95\% CI)}} \\
\midrule
ROC-AUC [$\uparrow$] & 0.47 (0.31--0.62) & 0.59 (0.43--0.75)* & 0.66 (0.51--0.81)* & \textbf{0.73 (0.58--0.86)*} \\
Accuracy [$\uparrow$] & 0.52 (0.39--0.63) & \textbf{0.67 (0.54--0.80)*} & 0.59 (0.46--0.72)* & \textbf{0.67 (0.54--0.80)*} \\
Precision [$\uparrow$] & 0.62 (0.45--0.78) & 0.67 (0.53--0.81) & \textbf{0.83 (0.65--1.00)*} & 0.81 (0.64--0.95)* \\
Recall [$\uparrow$] & 0.59 (0.41--0.74) & \textbf{0.91 (0.81--1.00)*} & 0.44 (0.28--0.61) & 0.62 (0.45--0.78)* \\
F1 Score [$\uparrow$] & 0.61 (0.45--0.72) & \textbf{0.78 (0.67--0.87)*} & 0.58 (0.40--0.72) & 0.70 (0.56--0.82)* \\
\bottomrule
\end{tabular}%
}%
}

\end{table}

We conducted an ablation study to quantify the impact of the main architectural and training choices on model performance. Specifically, we compared four configurations: the proposed multiloss with LoRA fine tuning (ML-FT), a cross-entropy only variant obtained by removing the contrastive loss from the previous model (CE-FT), a multiloss without fine tuning in which the backbone is fully frozen (ML-FR), and the baseline model that corresponds to our re-implementation of Yin et al.\cite{yin2023predicting} architecture. All models were evaluated on the same test set and with the same bootstrapping procedure. Quantitative performance measures are summarized in Table~\ref{tab:table_performance}.

From an interpretative perspective, introducing the contrastive loss improves the discriminative ability compared to CE alone (ML-FT: AUC 0.73 vs CE-FT: AUC 0.65), consistent with the hypothesis that explicitly supervising the geometry of the embedding space promotes a more robust separation between responders and non-responders. The fine tuning ablation shows a marked impact on AUC (ML-FT: AUC 0.73 vs ML-FR: AUC 0.58), suggesting that parameter-efficient fine tuning, even when limited to the last backbone block, contributes substantially to domain transfer toward mask lesion inputs. At the same time, the configuration without fine tuning ML-FR exhibits very high recall (0.91) and thus a higher F1 (0.77), but with lower precision (0.67) and, most importantly, a lower AUC: this profile is compatible with a decision behavior more oriented toward predicting the positive class, which can increase true positives but reduce the overall separation between the scores of the two classes. Finally, ML-FT achieved a higher ROC-AUC than the reproduced Yin et al. baseline when both were evaluated within the same pipeline and on the same test set (ML-FT:0.73 vs CE-FT: 0.47), indicating stronger discriminative performance of our approach under these experimental conditions.
All ROC-AUC differences between the ML-FT model and the other models were statistically significant according to DeLong’s test.

\subsection{Confidence-aware error analysis}

Confidence-aware error analysis consisted of manually inspecting uncertain correct cases (correct but low-confidence predictions) and confident wrong cases (incorrect but high-confidence predictions), revealing that many uncertain correct predictions lie close to the intrinsic RECIST 1.1 cutoff used to define a patient as responder or non-responder, indicating that some borderline decisions reflect ambiguity in the reference standard rather than purely model uncertainty.
Meanwhile, confident wrong cases cannot be explained solely by proximity to the RECIST 1.1 cutoff and were treated as potential failure modes: cases in which the model produces a clear but incorrect decision, warranting targeted review to identify possible systematic factors (e.g., acquisition noise).
\subsection{Calibration Analysis}
Calibration was assessed via a reliability diagram, which shows the calibration curve predominantly above the perfect calibration line: for mean predicted probabilities of 0.13 (n=9), 0.31 (n=18), and 0.49 (n=12), the empirical fraction of positives is 0.33, 0.50, and 0.75, respectively; similarly, at higher-confidence intervals the observed values are 0.74 vs 0.75 (n=8) and 0.87 vs 1.00 (n=7). Here, n denotes the number of test samples falling within each predicted probability interval. Overall, the systematic tendency of the observed positive fractions to exceed the corresponding mean predicted probabilities indicates that, on the test set, the model produces under-confident estimates for the positive class.
\subsection{Analysis of attention}
Attention weights quantify the relative contribution of each axial slice to the aggregated volumetric representation and, consequently, to the final prediction. In Figure 2, weights are visualized with a color map, where cooler tones denote lower contribution and warmer tones denote higher contribution. Figure 2a shows a correctly classified example, where higher weights are predominantly assigned to slices intersecting the lesion core, whereas slices near the superior and inferior boundaries receive lower weights. This pattern aligns with the intuition that central slices, typically containing a larger tumor cross-section, convey richer and more discriminative morphological information than peripheral slices. Overall, this qualitative observation suggests that the attention module learns a meaningful spatial prior, prioritizing the most informative regions along the axial extent.
In the misclassified example (Figure 2b), attention appears more diffuse and less selectively concentrated on the lesion core, indicating that the central-slice emphasis observed in correct predictions may be reduced in harder cases.
\begin{figure}[t]
    \centering
    \begin{subfigure}[b]{0.37\textwidth}
        \centering
        \includegraphics[width=\textwidth]{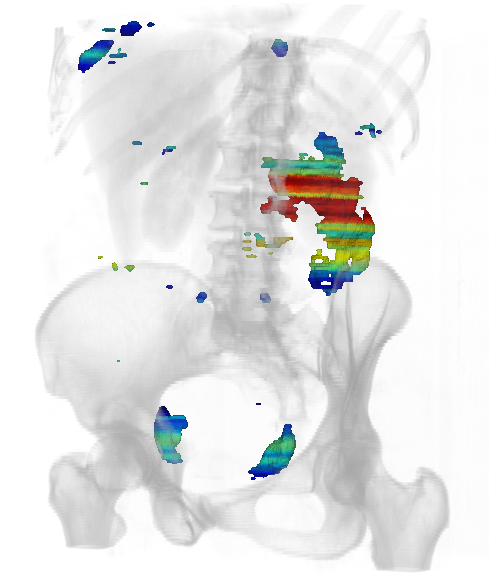}
        \caption{Correctly classified}
        \label{fig:attention-correct}
    \end{subfigure}%
    \hfill%
    \begin{subfigure}[b]{0.3\textwidth}
        \centering
        \includegraphics[width=\textwidth]{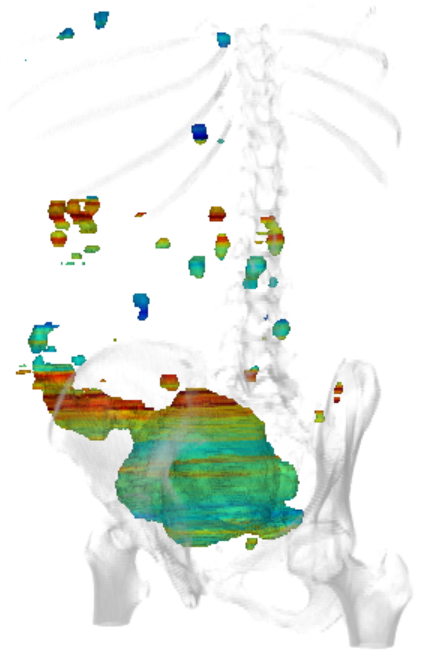}
        \caption{Misclassified}
        \label{fig:attention-misclassified}
    \end{subfigure}
    \caption{Attention-weight visualization across axial slices for two cases. Colors encode slice contribution (cooler: lower, warmer: higher). (a) Correct classification with weights concentrated on the lesion core. (b) Misclassification with a more diffuse attention distribution.}
    \label{fig:attention-main}
\end{figure}

\section{Discussion and Conclusion}
In this work, we developed a non-invasive deep learning framework to predict response to NACT from pre-treatment 3D lesion masks, with the goal of anticipating non-responder patients before therapy initiation. The proposed multi-loss model with LoRA adaptation achieved the best overall performance on the test set, reaching a ROC-AUC of 0.73 (95\% CI [0.58, 0.85]) with an accuracy of 0.66 and an F1-score of 0.70. Ablation results support the central design choices of the approach: the supervised contrastive margin term improves global separability over a cross-entropy-only variant and the evaluated state-of-the-art baseline. Qualitative analysis of the slice-attention suggests that the model aggregates volumetric information coherently, prioritizing the central lesion region. Beyond discrimination, calibration analysis highlighted a systematic under-confidence for the positive class, this suggests that the model’s probabilities are conservative rather than over-optimistic, which is desirable in a decision-support setting and can be further improved through post-hoc calibration. Finally, a confidence-aware error analysis showed that several uncertain-but-correct predictions concentrate near the intrinsic RECIST cutoff used to define responders versus non-responders, indicating that a meaningful fraction of ambiguity is driven by borderline ground truth rather than unstable modeling; in contrast, confident wrong cases emerge as actionable failure modes that can guide targeted review and future refinements. Overall, these findings strongly support the main objective of the work: using only pre-treatment lesion morphology, the proposed architecture learns clinically relevant predictive patterns and provides a robust foundation for an imaging-based stratification tool capable of anticipating NACT response, with clear potential to reduce ineffective therapy and avoid delays in optimal surgical management.
\\
\\
\textbf{Conflict of Interest} All authors have no conflicts of interest.
\\
\\
\textbf{Acknowledgements}
The work is part of the project \textit{Under-XAI: understanding ovarian cancer initiation and progression through explainable AI}, which received funding from the National Recovery and Resilience Plan (PNRR), Mission 6– Health, as part of the initiative \textit{Strengthening and Enhancement of Biomedical Research of the National Health Service}, under the EU’s NextGenerationEU program.
Project code: PNRR-MAD-2022-12376574. CUP: J47G22000530001.

\bibliographystyle{splncs04}
\bibliography{bibliography.bib}

@article{stat4,
    title = {Cancer stat facts: ovarian cancer. Available at: seer.cancer.gov/statfacts/html/ovary.html. [Accessed october 2025]},
    journal = {Surveillance, Epidemiology, and End Results Program.},
    year = {2025},
    publisher = {National Cancer Institute.},
    
}

@article{eisenhauer2009new,
  title={New response evaluation criteria in solid tumours: revised RECIST guideline (version 1.1)},
  author={Eisenhauer, Elizabeth A and Therasse, Patrick and Bogaerts, Jan and Schwartz, Lawrence H and Sargent, Danielle and Ford, Robert and Dancey, Janet and Arbuck, Stephen and Gwyther, Steve and Mooney, Margaret and others},
  journal={European journal of cancer},
  volume={45},
  number={2},
  pages={228--247},
  year={2009},
  publisher={Elsevier}
}

@article{crispin2023integrated,
  title={Integrated radiogenomics models predict response to neoadjuvant chemotherapy in high grade serous ovarian cancer},
  author={Crispin-Ortuzar, Mireia and Woitek, Ramona and Reinius, Marika AV and Moore, Elizabeth and Beer, Lucian and Bura, Vlad and Rundo, Leonardo and McCague, Cathal and Ursprung, Stephan and Escudero Sanchez, Lorena and others},
  journal={Nature communications},
  volume={14},
  number={1},
  pages={6756},
  year={2023},
  publisher={Nature Publishing Group UK London}
}

@article{yin2023predicting,
  title={Predicting neoadjuvant chemotherapy response and high-grade serous ovarian cancer from CT images in ovarian cancer with multitask deep learning: a multicenter study},
  author={Yin, Rui and Guo, Yijun and Wang, Yanyan and Zhang, Qian and Dou, Zhaoxiang and Wang, Yigeng and Qi, Lisha and Chen, Ying and Zhang, Chao and Li, Huiyang and others},
  journal={Academic Radiology},
  volume={30},
  pages={S192--S201},
  year={2023},
  publisher={Elsevier}
}

@article{fati2025deep,
  title={Deep Learning for Decision Support in Ovarian Cancer Treatment Planning},
  author={Fati, Francesca and Rosanu, Marina and De Vitis, Luigi and Rota, Alberto and Traversa, Alice and Ribero, Lucia and Schivardi, Gabriella and Petralia, Giuseppe and Aletti, Giovanni Damiano and Colombo, Nicoletta and others},
  year={2025}
}

@article{gupta2024margin,
  title={Margin-aware optimized contrastive learning for enhanced self-supervised histopathological image classification},
  author={Gupta, Ekta and Gupta, Varun},
  journal={Health Information Science and Systems},
  volume={13},
  number={1},
  pages={2},
  year={2024},
  publisher={Springer}
}

@article{buddenkotte2023deep,
  title={Deep learning-based segmentation of multisite disease in ovarian cancer},
  author={Buddenkotte, Thomas and Rundo, Leonardo and Woitek, Ramona and Escudero Sanchez, Lorena and Beer, Lucian and Crispin-Ortuzar, Mireia and Etmann, Christian and Mukherjee, Subhadip and Bura, Vlad and McCague, Cathal and others},
  journal={European radiology experimental},
  volume={7},
  number={1},
  pages={77},
  year={2023},
  publisher={Springer}
}

@article{simeoni2025dinov3,
  title={Dinov3},
  author={Sim{\'e}oni, Oriane and Vo, Huy V and Seitzer, Maximilian and Baldassarre, Federico and Oquab, Maxime and Jose, Cijo and Khalidov, Vasil and Szafraniec, Marc and Yi, Seungeun and Ramamonjisoa, Micha{\"e}l and others},
  journal={arXiv preprint arXiv:2508.10104},
  year={2025}
}

@article{hu2022lora,
  title={Lora: Low-rank adaptation of large language models.},
  author={Hu, Edward J and Shen, Yelong and Wallis, Phillip and Allen-Zhu, Zeyuan and Li, Yuanzhi and Wang, Shean and Wang, Liang and Chen, Weizhu and others},
  journal={Iclr},
  volume={1},
  number={2},
  pages={3},
  year={2022}
}

@article{intro0,
  title={High-grade serous ovarian cancer: basic sciences, clinical and therapeutic standpoints},
  author={Lisio, Michael-Antony and Fu, Lili and Goyeneche, Alicia and Gao, Zu-hua and Telleria, Carlos},
  journal={International journal of molecular sciences},
  volume={20},
  number={4},
  pages={952},
  year={2019},
  publisher={MDPI}
}

@article{intro8,
  title={Management of newly diagnosed or recurrent ovarian cancer.},
  author={Matulonis, Ursula A},
  journal={Clinical Advances in Hematology and Oncology},
  year={2018}
}

\end{document}